\documentclass[sigconf]{acmart}

\AtBeginDocument{%
  \providecommand\BibTeX{{%
    \normalfont B\kern-0.5em{\scshape i\kern-0.25em b}\kern-0.8em\TeX}}}

\copyrightyear{2022}
\acmYear{2022}
\setcopyright{acmcopyright}
\acmConference[MM '22] {Proceedings of the 30th ACM International Conference on Multimedia }{October 10--14, 2022}{Lisboa, Portugal.}
\acmBooktitle{Proceedings of the 30th ACM International Conference on Multimedia (MM '22), October 10--14, 2022, Lisboa, Portugal}
\acmPrice{15.00}
\acmISBN{978-1-4503-9203-7/22/10}
\acmDOI{10.1145/3503161.3548389}

\usepackage{subfigure}
\usepackage{multirow}
\usepackage{color}
\usepackage{hyperref}
\usepackage[normalem]{ulem}
\useunder{\uline}{\ul}{}
\usepackage{balance}
\usepackage{cases}
\usepackage{array}
\usepackage{bbding}

\begin{document}
\title{Fast Hierarchical Deep Unfolding Network for \\ Image Compressed Sensing}

\author{Wenxue Cui}
\affiliation{%
  \institution{Harbin Institute of Technology}
  \city{Harbin}
  \country{China}}
\email{wenxuecui@stu.hit.edu.cn}

\author{Shaohui Liu}
\affiliation{%
  \institution{Harbin Institute of Technology}
  \city{Harbin}
  \country{China}}
\affiliation{%
  \institution{Peng Cheng Lab}
  \city{Shenzhen}
  \country{China}
}

\author{Debin Zhao}
\affiliation{%
  \institution{Harbin Institute of Technology}
  \city{Harbin}
  \country{China}}
\affiliation{%
  \institution{Peng Cheng Lab}
  \city{Shenzhen}
  \country{China}
}

\renewcommand{\shortauthors}{Wenxue Cui et al.}

\begin{abstract}
  By integrating certain optimization solvers with deep neural network, deep unfolding network (DUN) has attracted much attention in recent years for image compressed sensing (CS). However, there still exist several issues in existing DUNs: 1) For each iteration, a simple stacked convolutional network is usually adopted, which apparently limits the expressiveness of these models. 2) Once the training is completed, most hyperparameters of existing DUNs are fixed for any input content, which significantly weakens their adaptability. In this paper, by unfolding the Fast Iterative Shrinkage-Thresholding Algorithm (FISTA), a novel fast hierarchical DUN, dubbed FHDUN, is proposed for image compressed sensing, in which a well-designed hierarchical unfolding architecture is developed to cooperatively explore richer contextual prior information in multi-scale spaces. To further enhance the adaptability, series of hyperparametric generation networks are developed in our framework to dynamically produce the corresponding optimal hyperparameters according to the input content. Furthermore, due to the accelerated policy in FISTA, the newly embedded acceleration module makes the proposed FHDUN save more than 50\% of the iterative loops against recent DUNs. Extensive CS experiments manifest that the proposed FHDUN outperforms existing state-of-the-art CS methods, while maintaining fewer iterations.

\end{abstract} 



\maketitle

\section{Introduction}
Compressed sensing (CS)\cite{donoho2006compressed, candes2008introduction} theory demonstrates that if a signal is sparse in a certain domain, it can be recovered with a high probability from a much fewer acquired measurement than prescribed by the Nyquist sampling theorem. Recently, the benefits of its reduced sampling rate have attracted many practical applications, including but not limited to single-pixel imaging\cite{4472247,7778203}, magnetic resonance imaging (MRI)\cite{Lustig2008Compressed,8417964} and snapshot compressive imaging\cite{8481592,9010044}.

Mathematically, given the input signal $x\in \mathbb{R}^{N}$, the sampled linear measurements $y\in \mathbb{R}^{M}$ can be obtained by $y=\Phi x$, where $\Phi \in \mathbb{R}^{M\times N}$ is the sampling matrix with $M\ll N$ and $\frac{M}{N}$ is the CS sampling ratio. Obviously, the signal recovery from the compressed measurements is to solve an under-determined linear inverse problem\cite{8765626}, and the corresponding optimization model can be formulated as follows:
\vspace{-0.05in}
\begin{equation}
\tilde{x}=\mathop{\arg}\mathop{\min}_{x}\frac{1}{2} \| \Phi x-y \|_{2}^{2}+\lambda\psi(x)
\label{Eq:111}
\vspace{-0.04in}
\end{equation}
where $\psi(x)$ is the prior term with the regularization parameter $\lambda$. In the traditional CS methods\cite{gao2015block,Kim2010Compressed,Metzler2016From,6190204}, the prior term can be the sparsifying operator in certain pre-defined transform domains (such as DCT\cite{6890254} and wavelet\cite{7122281}). To further enhance the reconstructed quality, more sophisticated structures are established, including minimal total variation\cite{1580791,li2013tval3}, low rank\cite{6827224,6288484} and non-local self-similarity image prior\cite{9190055,zhang2014group}. Many of these approaches have led to significant improvements. However, these optimization-based CS reconstruction algorithms usually suffer from high computational complexity because of their hundreds of iterations, thus limiting the practical applications of CS greatly.

Recently, fueled by the powerful learning ability of deep neural networks, many deep network-based image CS methods have been proposed. According to the intensity of interpretability, the existing CS networks can be roughly grouped into the following two categories: Uninterpretable deep black box CS network (DBN) and interpretable deep unfolding CS network (DUN). \textbf{1) Uninterpretable DBN:} this kind of method\cite{Yao2017DR2,7780424,cui2018efficient,8765626,shi2019scalable} usually trains the deep network as a black box, and builds a direct deep inverse mapping from the measurement domain to the original signal domain. Due to the simplicity of such kind of algorithm, it has been widely favored in the early stage of deep CS research. Unfortunately, this rude mapping strategy makes this type of method lack essential theoretical basis and interpretability, thus limiting the reconstructed quality significantly. \textbf{2) Interpretable DUN:} this kind of method\cite{2017Learned,8578294,9019857,9298950,9467810} usually unfolds certain optimization algorithms, such as iterative shrinkage-thresholding algorithm (ISTA)\cite{DBLP-ISTA} and approximate message passing (AMP)\cite{donoho2009message}, into the forms of deep network, and learns a truncated unfolding function by an end-to-end fashion. Inspired by the iterative mode of the optimization method, DUN is usually composed of a fixed number of stages (corresponding to the iterations of the optimization algorithms) to gradually reconstruct the target signal, which undoubtedly makes it enjoy solid theoretical support and better interpretability.

Compared to DBN, the recent DUN has become the mainstream for CS reconstruction. However, there still exist several burning issues in existing DUNs: 1) For each iteration, a simple stacked convolutional network is usually adopted, which weakens the perception of wider context information that limits the expressiveness of these models for image reconstruction. 2) Once the training is completed, most hyper-parameters (e.g., the step size\cite{9019857} and the control parameter\cite{9298950}) of existing DUNs are fixed for any input content, which weakens the adaptive ability of these models. 


To overcome above issues, we propose a novel Fast Hierarchical Deep Unfolding Network (FHDUN) for image compressed sensing by unfolding FISTA. In the proposed framework, a well-designed hierarchical unfolding architecture, as shown in Fig.1, is developed, which can cooperatively infer the solver FISTA in multi-scale spaces to perceive wider contextual prior information. Due to the hierarchical design, the proposed framework is able to persist and transmit richer textures and structures among cascading iterations for boosting reconstructed quality. In addition, series of hyperparametric generation networks are developed in our framework to dynamically produce the corresponding optimal hyperparameters according to the input content, which enhances the adaptability of the proposed model significantly. Moreover, by incorporating FISTA's acceleration strategy, the proposed FHDUN saves more than 50\% of the iterative loops against recent DUNs. Extensive experiments demonstrate that the proposed FHDUN achieves better reconstructed performance against recent DUNs with much fewer iterations.

\begin{figure*}
\centering
\includegraphics[width=\textwidth]{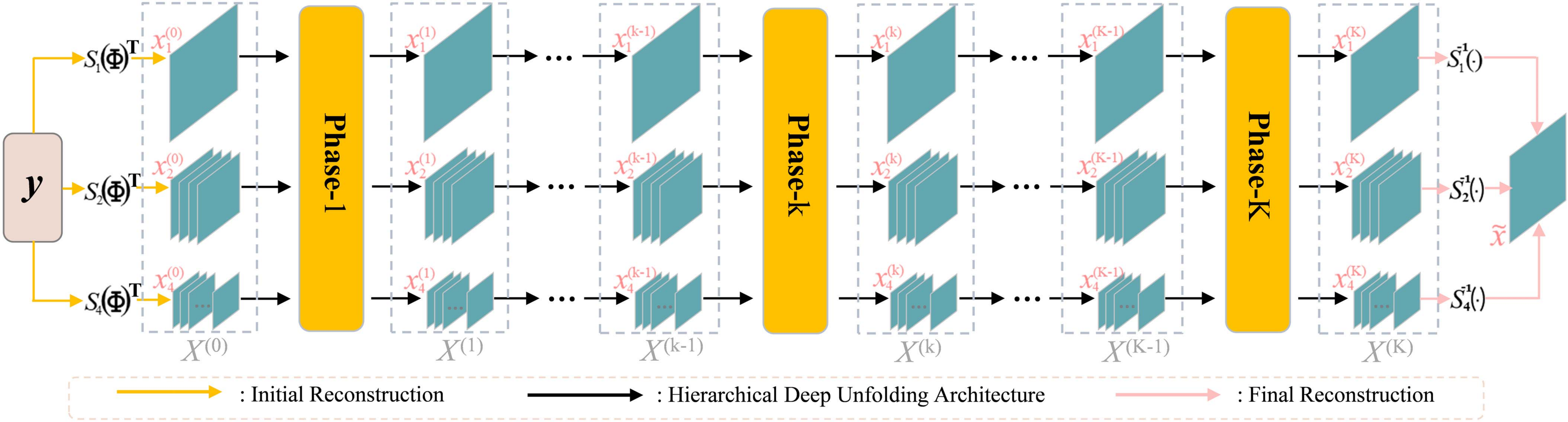}
\vspace{-0.26in}\caption{Diagram of the proposed FHDUN, which consists of multiple phases to gradually reconstruct the target image. In FHDUN, series of branches of different scales are included to form a hierarchical structure. }
\vspace{-0.12in}
\label{Fig:2}
\end{figure*}

The main contributions are summarized as follows:

\textbf{(1)} A novel Fast Hierarchical Deep Unfolding Network (FHDUN) for image CS is proposed, which can cooperatively infer the optimization solver FISTA in multi-scale spaces to explore richer contextual prior information. 

\textbf{(2)} In our framework, series of hyperparametric generation networks are designed to dynamically generate the optimal hyperparameters according to the input content, which greatly enhances the adaptability of the proposed model.

\textbf{(3)} Due to the introduction of the acceleration policy by unfolding the traditional FISTA, the proposed FHDUN is able to save more than 50\% of the iterative loops compared with the recent deep unfolding CS networks.

\textbf{(4)} Extensive experiments show that the proposed FHDUN outperforms existing state-of-the-art CS reconstruction networks by large margins with fewer iterations.

\section{Related Work}
\subsection{Uninterpretable DBN}
Deep black box network (DBN) usually directly builds an inverse deep mapping from the measurement domain to the original image domain. Due to its simplicity, this kind of method is widely favored by many researchers. Specifically, the early works\cite{7780424,2018Convolutional,2019DR2} usually reconstruct the target image block-by-block and then splice the reconstructed image blocks together into a final reconstructed image. However, these block-by-block methods usually suffer from serious block artifacts\cite{cui2018efficient}. To relieve this problem, several literatures\cite{shi2019scalable,8765626,8019428,9199540} attempt to explore the deep image priors in the whole image space. Specifically, these CS methods still adopt the block-based sampling\cite{gan2007block}, however during the reconstruction, they first concatenate all image blocks together in the initial reconstruction, and then complete a deep reconstruction in the whole image space. More recently, to enhance the applicability of CS framework, several scalable CS networks\cite{shi2019scalable,xu2018lapran} are proposed, which achieve scalable sampling and reconstruction with only one model. 

Compared with the traditional optimization-based CS methods, the aforementioned deep black box CS networks can automatically explore image priors on massive training data and achieve higher reconstruction performance with fast computational speed. However, these CS networks usually train the deep network as a black box, which makes these methods lack essential theoretical basis and interpretability, thus limiting the reconstructed quality significantly. 

\subsection{Interpretable DUN}
Deep unfolding network (DUN) usually unfolds certain optimization algorithms into deep network forms to enjoy a better interpretability, which has been applied to solve diverse low-level vision problems, such as denoising\cite{8100106}, debluring\cite{8237753}, and demosaicking \cite{eccvdemosaicking}. Recently, considering CS reconstruction, some DUNs attempt to integrate some effective convolutional neural networks with some optimization methods including half quadratic splitting (HQS)\cite{8434321,8099783}, alternating minimization, approximate message passing (AMP)\cite{9298950,9159912} and alternating direction method of multipliers (ADMM)\cite{8550778}. As above, different optimization algorithms usually lead to different optimization-inspired DUNs.

Iterative shrinkage-shresholding algorithm (ISTA)\cite{DBLP-ISTA}, as a prevailing optimization method, has been widely used to solve many large-scale linear inverse problems. To solve Eq.(1), each iteration of ISTA involves gradient descent updating followed by a proximal operator:
\begin{equation}
x^{(k)} = \mathcal{H}(x^{(k-1)} - \rho\Phi^{\rm T}(\Phi x^{(k-1)} - y))
\label{Eq:1}
\vspace{-0.02in}
\end{equation}
where $\rho$ is the step size for gradient descent updating and $\mathcal{H}$ indicates a specific shrinkage/soft-threshold function (i.e., proximal operator). Recently, several excellent DUNs attempt to embed deep networks into ISTA to solve CS problem by iterating between the following update steps:
\vspace{-0.01in}
\begin{numcases}{}
r^{(k)} = x^{(k-1)} - \rho^{(k)}\Phi^{\rm T}(\Phi x^{(k-1)}-y) \\
x^{(k)} = \mathcal{H}_{\psi}^{(k)}(r^{(k)})
\label{eq:1}
\end{numcases}
where Eq.(3) is responsible for the gradient descent of linear projection and $\rho^{(k)}$ is a learnable step size. Eq.(4) corresponds to a specific proximal operator, which is usually fitted by series of convolutional layers to learn a deep proximal mapping. $\mathcal{H}_{\psi}^{(k)}$ signifies the embedded deep network for exploring the image prior $\psi$. Based on Eqs.(3) and (4), several ISTA-inspired CS DUNs\cite{8578294,9019857,2021Memory,9467810} have been proposed to reconstruct the target image from measurements. 

Apparently, compared with deep black box CS network, the aforementioned deep unfolding CS networks have better interpretability. However, these algorithms usually adopt a plain network architecture and therefore cannot fully exert the expressiveness of the proposed model for image reconstruction. Besides, due to the absence of the acceleration strategy in recent DUNs, the convergence usually requires dozens of iterations, for example MADUN\cite{2021Memory} and COAST\cite{9467810} respectively need 25 and 20 iterations, which limits their practical applications in some real-time CS systems.

\section{Proposed Method}

\begin{figure}[b]
\centering
\vspace{-0.15in}
\includegraphics[width=\linewidth]{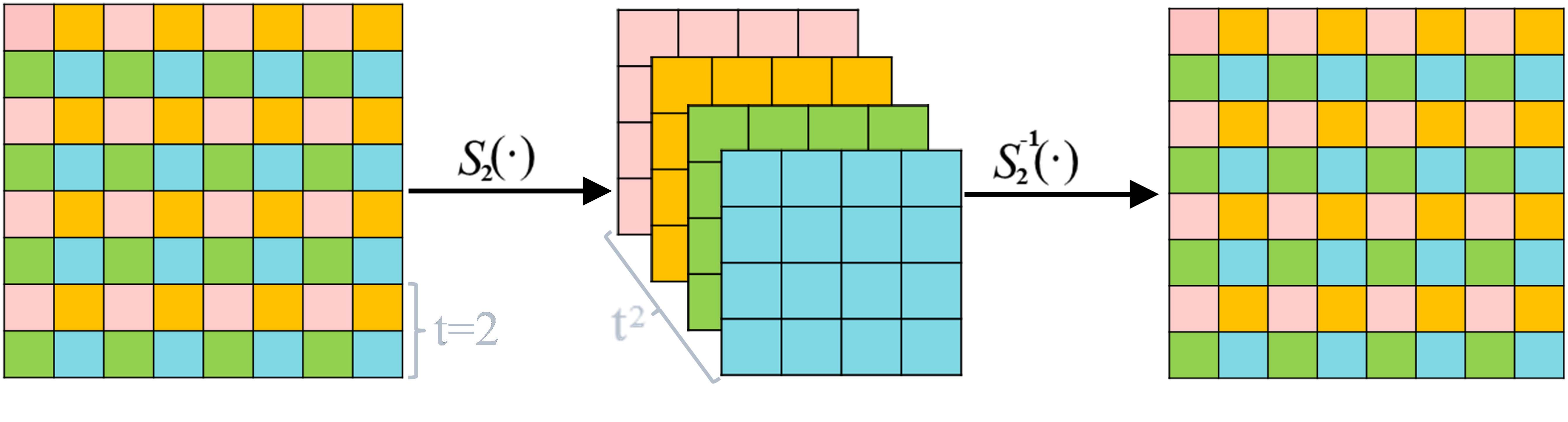}
\vspace{-0.28in}\caption{The operation details of the introduced unshuffle operator $S_{t}$ and its inverse operator $S_{t}^{-1}$ under the condition of scale factor $t$$=$2.}
\label{Fig:4}
\end{figure}

\begin{figure*}
\centering
\includegraphics[width=\textwidth]{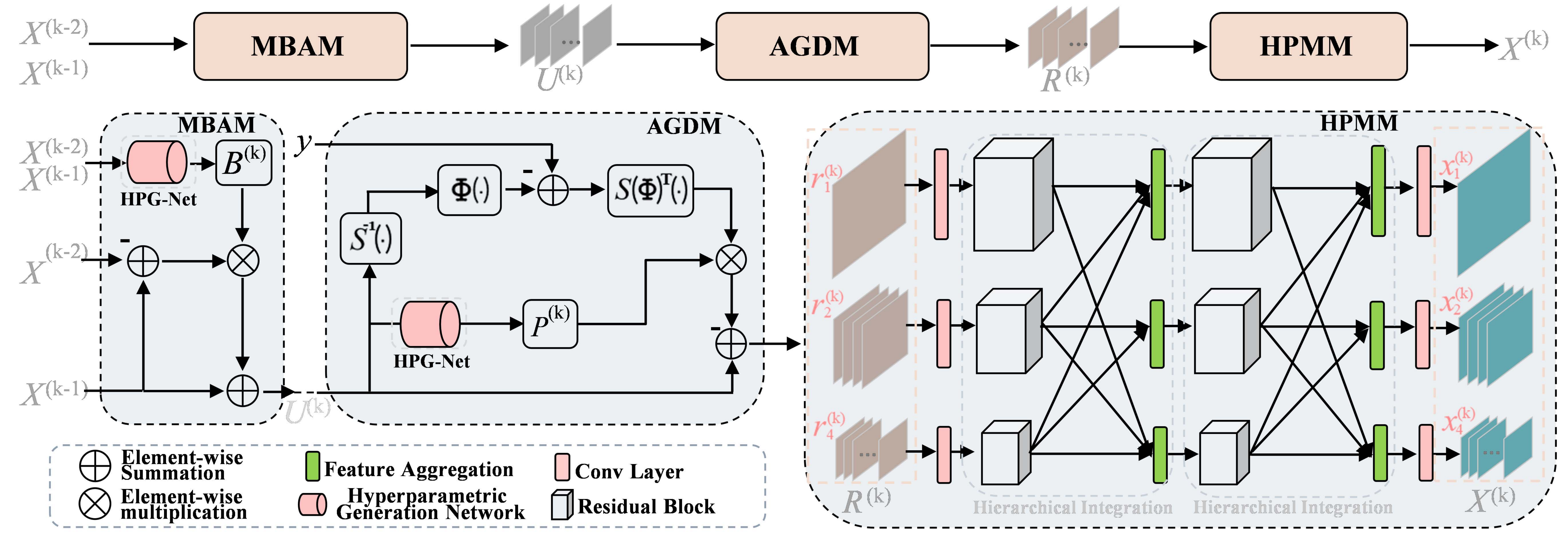}
\vspace{-0.24in}\caption{The network in each phase, which includes three modules: Momentum Based Acceleration Module (MBAM), Adaptive Gradient Descent Module (AGDM), Hierarchical Proximal Mapping Module (HPMM).}
\vspace{-0.13in}
\label{Fig:5}
\end{figure*}

\subsection{Preliminary}
As described in Subsection 2.2, ISTA provides rich theoretical inspirations for many recent deep unfolding CS networks. However, ISTA is generally recognized as a time-consuming method because of its requirement for tedious repeated iterations. To alleviate this problem, two fast versions of ISTA are proposed, including the two-step IST (TwIST) algorithm\cite{4358846} and fast IST algorithm (FISTA)\cite{DBLP-ISTA}. By embedding the momentum-based acceleration method into ISTA, FISTA solves Eq.(1) by iterating the following update steps:
\vspace{-0.01in}
\begin{numcases}{}
t^{(k)} = \frac{1+\sqrt{1+4(t^{(k-1)})^{2}}}{2} \\
u^{(k)} = x^{(k-1)} + (\frac{t^{(k-1)}-1}{t^{(k)}})(x^{(k-1)}-x^{(k-2)}) \\
x^{(k)} = \mathcal{H}(u^{(k)}-\rho\Phi^{\rm T}(\Phi u^{(k)}-y))
\label{eq:2}
\end{numcases}
Compared with ISTA solver as shown in Eq.(2), the main improvement of FISTA is that the proximal operator $\mathcal{H}$ is not applied on the previous estimation $x_{(k-1)}$, but rather at $u_{(k)}$ which adopts a well-designed linear combination of the previous two estimations $x_{(k-1)}$ and $x_{(k-2)}$.

According to Eqs.(5)-(7), FISTA can be appropriately unfolded into the following iterative update steps:
\begin{numcases}{}
u^{(k)} = x^{(k-1)} + \beta^{(k)}(x^{(k-1)} - x^{(k-2)}) \\
r^{(k)} = u^{(k)} - \rho^{(k)} \Phi^{\rm T}(\Phi u^{(k)}-y) \\
x^{(k)} = \mathcal{H}_{\psi}^{(k)}(r^{(k)})
\label{eq:3}
\end{numcases}
where Eq.(8) is mainly responsible for acceleration by computing a new intermediate variable from $x_{(k-1)}$ and $x_{(k-2)}$, and $\beta^{(k)}$ represents the scalar for momentum update. Eqs.(9) and (10) have the same meaning as Eqs.(3) and (4), which respectively signify gradient descent and proximal mapping.

\subsection{Overview of FHDUN}

It is clear from Eqs.(8)-(10) that the variable $x^{(k)}$ bridges the gap between different iterations, and a rough $x^{(k)}$ usually hinders information transmission between adjacent iterations, thus losing much more image details during reconstruction. Based on above, we propose a novel hierarchical unfolding architecture (in Fig.1), which consists of multiple branches of different scales that form a hierarchical structure.

In order to build the hierarchical model integrating multiple scales, we first introduce a series of unshuffle operators $\{S_{t}\}$, which approximately can be regarded as the inverse operation of PixelShuffle (widely used in super-resolution tasks)\cite{7780576}, and $t$ has a similar meaning to the scale factor in PixelShuffle. Given $\{S_{t}\}$, we use $\{S^{-1}_{t}\}$ to represent their corresponding inverse operations. Fig.2 shows more details of the operators $S_{t}$ and $S^{-1}_{t}$ under the condition of $t$=2. As above, the update steps of Eqs.(8)-(10) in a specific scale space can be intuitively expressed as:
\begin{numcases}{}
u_{t}^{(k)} = x_{t}^{(k-1)} + \beta_{t}^{(k)}(x_{t}^{(k-1)} - x_{t}^{(k-2)}) \\
r_{t}^{(k)} = u_{t}^{(k)} - \rho_{t}^{(k)} S_{t}(\Phi)^{\rm T}(\Phi S_{t}^{-1}(u_{t}^{(k)})-y) \\
x_{t}^{\hspace{-0.02in}(k)} = \mathcal{H}_{\psi_{t}}^{(k)}(r_{t}^{(k)})
\label{eq:4}
\end{numcases}
where $x_{t}^{(k-1)}$, $x_{t}^{(k-2)}$, $u_{t}^{(k)}$, $r_{t}^{(k)}$ and $x_{t}^{(k)}$ are the variables under the scale factor of $t$. $\beta_{t}^{(k)}$ and $\rho_{t}^{(k)}$ are the corresponding learnable parameters. $\mathcal{H}_{\psi_{t}}^{(k)}$ indicates a deep neural network (proximal mapping) in current scale space (scale factor is $t$) to explore the prior knowledge $\psi_{t}$.

In order to collaboratively integrate the update steps (i.e., Eqs.(11)-(13)) in different scale spaces, we attempt to aggregate the unfolding models of different scales together and propose a novel hierarchical unfolding architecture:
\begin{numcases}{}
U^{(k)} = X^{(k-1)} + B^{(k)}(X^{(k-1)} - X^{(k-2)}) \\
R^{(k)} = U^{(k)} - P^{(k)} S(\Phi)^{\rm T}(\Phi S^{-1}(U^{(k)})-y) \\
X^{\hspace{-0.02in}(k)} = \mathcal{H}_{\Psi}^{(k)}(R^{(k)})
\label{eq:5}
\end{numcases}
where $X^{(k-1)}$, $X^{(k-2)}$, $U^{(k)}$, $R^{(k)}$ and $X^{(k)}$ are variable sets. For example, $X^{(k)}$=$\{x_{t}^{(k)}\}$, and $t=\{1,2,...\}$. $B^{(k)}$ and $P^{(k)}$ are learnable parameter sets. Analogously, $S(\cdot)$ and $S^{-1}(\cdot)$ respectively signify the collections of the unshuffle operators and their inverse versions under multiple scale spaces. $\mathcal{H}_{\Psi}^{(k)}$ is a unified deep neural network (proximal mapping) to explore the priors $\Psi$ in multiple scale spaces. It is noted that compared with $\psi$ and $\psi_{t}$, $\Psi$ represents more sophisticated priors, including the image priors in each scale space, the correlation priors among different scale spaces and so on.

For the initial reconstruction of our FHDUN, we set $X^{(0)}$= $S\hspace{-0.02in}(\Phi)^{\rm\hspace{-0.02in} T}y$, after which the update steps as shown in Eqs.(14)-(16) are iteratively performed. Finally, we average the outputs of different branches to obtain the final reconstruction $\tilde{x}$.

\begin{figure}[b]
\centering
\vspace{-0.15in}
\includegraphics[width=\linewidth]{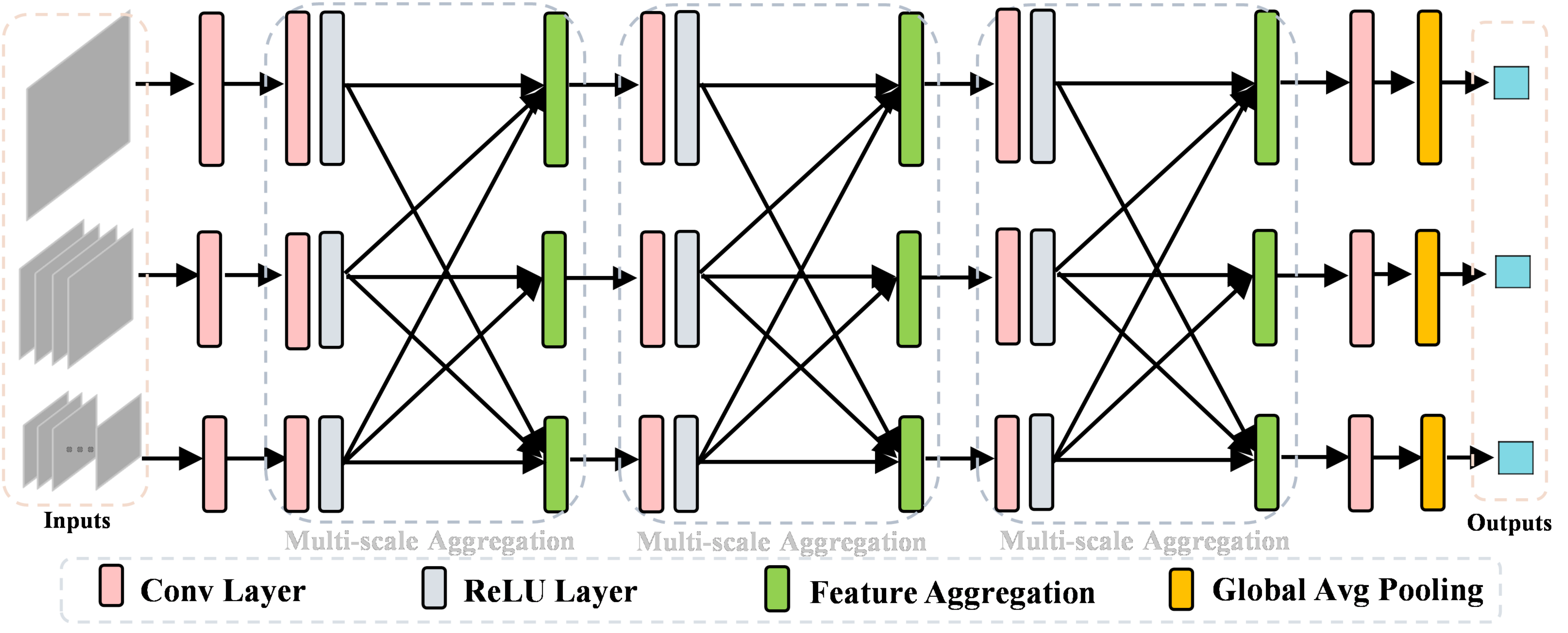}
\vspace{-0.25in}\caption{The network structure details of the proposed HPG-Net in MBAM and AGDM. The strcture of feature aggregation is shown in Fig.5(b).}
\label{Fig:1}
\end{figure}

Corresponding to Eqs.(14)-(16), each iteration of our FHDUN consists of three functional modules (in Fig.3): Momentum Based Acceleration Module (MBAM), Adaptive Gradient Descent Module (AGDM) and Hierarchical Proximal Mapping Module (HPMM). The following subsections will describe more details about these three modules.

\begin{table*}[t]
\centering
\caption{Average PSNR and SSIM comparisons of recent deep network-based CS algorithms using random sampling matrix on dataset Set11. Bold indicates the best result, underline signifies the second best result.}
\label{tab:1}
\vspace{-0.15in}
\small
\begin{tabular}{p{2.5cm}<{\centering} | p{0.86cm}<{\centering} p{0.91cm}<{\centering} | p{0.86cm}<{\centering}  p{0.91cm}<{\centering} | p{0.86cm}<{\centering} p{0.91cm}<{\centering} | p{0.86cm}<{\centering} p{0.91cm}<{\centering} | p{0.86cm}<{\centering} p{0.91cm}<{\centering} | p{0.86cm}<{\centering} p{0.91cm}<{\centering}}
\toprule
\multirow{2}*{Algorithms} & \multicolumn{2}{c}{Ratio=0.01} & \multicolumn{2}{c}{Ratio=0.10} & \multicolumn{2}{c}{Ratio=0.25} & \multicolumn{2}{c}{Ratio=0.30} & \multicolumn{2}{c}{Ratio=0.40} & \multicolumn{2}{c}{Avg.}\\
\cline{2-13}
&PSNR&SSIM&PSNR&SSIM&PSNR&SSIM&PSNR&SSIM&PSNR&SSIM&PSNR&SSIM\\
\midrule
ReconNet\cite{7780424}&17.54&0.4426&24.07&0.6958&26.38&0.7883&28.72&0.8517&30.59&0.8928&25.46&0.7342\\
I-Recon\cite{2018Convolutional}&\underline{19.80}&\underline{0.5018}&25.97&0.7888&28.52&0.8547&31.45&0.9135&32.26&0.9243&27.60&0.7966\\
DR$^{2}$-Net\cite{2019DR2}&17.44&0.4294&24.71&0.7175&--\ --&--\ --&30.52&--\ --&31.20&--\ --&--\ --&--\ --\\
DPA-Net\cite{9199540}&18.05&0.5011&26.99&0.8354&32.38&0.9311&33.35&0.9425&35.21&0.9580&29.20&0.8336\\
\hline
IRCNN\cite{8099783}&7.70&0.3324&23.05&0.6789&28.42&0.8382&29.55&0.8606&31.30&0.8898&24.00&0.7200\\
LDAMP\cite{2017Learned}&17.51&0.4409&24.94&0.7483&--\ --&--\ --&32.01&0.9144&34.07&0.9393&--\ --&--\ --\\
ISTA-Net$^{+}$\cite{8578294}&17.45&0.4131&26.49&0.8036&32.48&0.9242&33.81&0.9393&36.02&0.9579&29.25&0.8076\\
DPDNN\cite{8481558}&17.59&0.4459&26.23&0.7992&31.71&0.9153&33.16&0.9338&35.29&0.9534&28.80&0.8095\\
NN\cite{8878159}&17.67&0.4324&23.90&0.6927&29.20&0.8600&30.26&0.8833&32.31&0.9137&26.67&0.7564\\
MAC-Net\cite{eccvcs}&18.26&0.4566&27.68&0.8182&32.91&0.9244&33.96&0.9372&36.18&0.9562&29.80&0.8185\\
iPiano-Net\cite{SU2020115989}&19.38&0.4812&\underline{28.05}&\underline{0.8460}&\underline{33.53}&\underline{0.9359}&\underline{34.78}&\underline{0.9472}&\underline{37.00}&\underline{0.9631}&\underline{30.55}&\underline{0.8347}\\



\midrule

FHDUN&\textbf{20.18}&\textbf{0.5468}&\textbf{29.53}&\textbf{0.8859}&\textbf{35.01}&\textbf{0.9512}&\textbf{36.12}&\textbf{0.9589}&\textbf{38.04}&\textbf{0.9696}&\textbf{31.78}&\textbf{0.8625}\\
\bottomrule

\end{tabular}
\vspace{-0.1in}
\label{tab:2}
\end{table*}

\subsection{MBAM for Fast Convergence}
From Eq.(14), the mission of MBAM is to accelerate the convergence by computing a new variable set $U^{(k)}$ from $X^{(k-1)}$ and $X^{(k-2)}$. Apparently, the momentum scalar $B^{(k)}$ has a crucial influence on the convergence efficiency of the proposed model. Different from the existing hyperparametric learning strategies (once the training is completed, the learned hyperparameters are fixed for any input content), a novel hyperparametric generation network (HPG-Net) as shown in Fig.4 is designed in our framework to dynamically learn the momentum scalar $B^{(k)}$ from the previous iterations:
\vspace{-0.01in}
\begin{equation}
B^{(k)} = \{\beta_{t}^{(k)}\} = F_{(\theta_{u}^{k})}(X^{(k-1)}, X^{(k-2)})
\vspace{-0.0in}
\end{equation}
where $F_{(\theta_{u}^{k})}$ indicates the proposed HPG-Net in the current iteration with the learnable parameter $\theta_{u}^{k}$, and its input is the concatenated mixture of $X^{(k-1)}$ and $X^{(k-2)}$. Due to the hierarchical structure of the proposed framework, the developed HPG-Net is able to hierarchically integrate the multi-scale information from different branches. Specifically, three multi-scale aggregation submodules (shown in gray dashed boxes of Fig.4) are introduced in HPG-Net, which can cooperatively explore more information from different scale spaces. After the aggregation submodules, series of convolutional layers and global average pooling layers are appended to generate the final momentum scalars.

\begin{figure}[h]
\centering
\vspace{-0.05in}
\includegraphics[width=\linewidth]{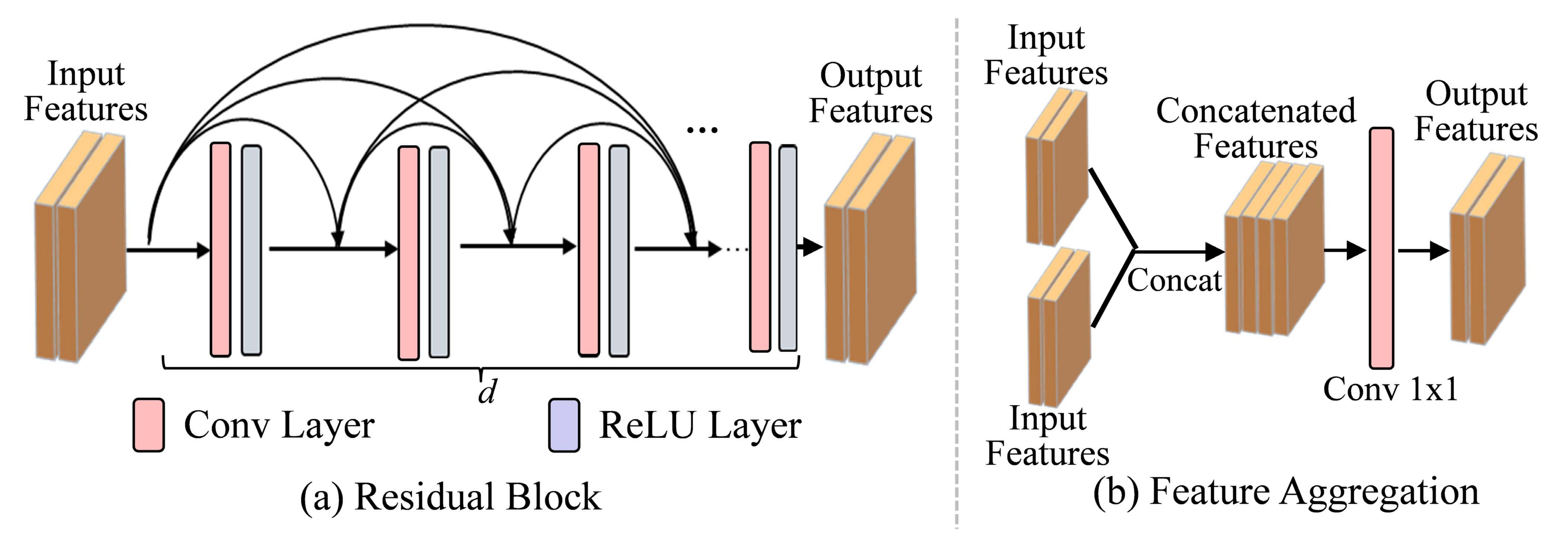}
\vspace{-0.32in}\caption{The network structure details of the residual block (a) and the feature aggregation (b).}
\vspace{-0.2in}
\label{Fig:3}
\end{figure}

\subsection{AGDM for Gradient Descent}
In Eq.(15), AGDM performs the gradient descent for the linear projection. Similar to $B^{(k)}$ (in Eq.(14)), the step size $P^{(k)}$ is also a crucial hyper-parameter that controls the magnitude of the gradient updating. According to Eqs.(3) and (4), the existing ISTA-inspired DUNs directly optimize the step size value from the training data, and once the training is completed, the step size of these methods is fixed for any input content. In our framework, similar to the learning strategy of momentum scalar $B^{(k)}$ (Subsection 3.3), another novel hyperparametric generation network (HPG-Net) is designed (also shown in Fig.4) to adaptively learn the corresponding optimal step size for the given input content:
\vspace{-0.01in}
\begin{equation}
P^{(k)} = \{\rho_{t}^{(k)}\} = J_{(\theta_{r}^{k})}(U^{(k)})
\vspace{-0.0in}
\end{equation}
where $J_{(\theta_{r}^{k})}$ indicates the network HPG-Net in current iteration with the learnable parameter $\theta_{r}^{k}$, and its input is the intermediate variable set $U^{(k)}$ from Eq.(14). It is noted that the proposed HPG-Net in AGDM has the similar network structure with that of MBAM, which can efficiently integrate the multi-scale information from different branches. For more details of HPG-Net, the multi-scale aggregation submodules (gray dashed boxes of Fig.4) are still retained, which can cooperatively explore more information from multiple scale spaces. Finally, series of convolutional layers (with single kernel) and global average pooling layers are appended to generate the final step sizes for different branches.

\subsection{HPMM for Proximal Mapping}
After gradient descent, the module HPMM corresponding to Eq.(16) is appended, which actually is a deep convolutional neural network (dubbed HPM-Net) in our framework. For more details of the proposed HPM-Net, a series of horizontal branches are included as shown in Fig.3 to form a hierarchical structure. Specifically, the horizontal branch is responsible for the feature extraction under a certain scale space. The transformations among different scales are performed between all the horizontal branches, including both downsampling and upsampling operations. For more internal details, two hierarchical aggregation submodules (the gray dashed boxes of HPMM in Fig.3) are introduced, in which series of residual blocks (Fig.5(a)) and feature aggregation submodules (Fig.5(b)) are utilized to capture and aggregate the deep features of different scales. For simplicity, the learnable parameter in HMP-Net of current iteration is represented as $\theta_{x}^{k}$, which will be utilized in the following sections.

Compared with the proximal mapping of existing DUNs (usually adopt a plain convolutional network to explore image priors at a single scale space), the proposed hierarchical network can cooperatively perceive wider contextual prior information in multi-scale spaces. In addition, due to the hierarchical architecture design, the proposed DUN can persist and transmit more richer information between neighbouring iterations for boosting reconstructed quality.

\begin{table*}[t]
\centering
\caption{Average PSNR and SSIM comparisons of recent deep network-based CS algorithms using learned sampling matrix on dataset Set5. Bold indicates the best result, and underline signifies the second best result.}
\label{tab:3}
\vspace{-0.15in}
\small
\begin{tabular}{p{3.5cm}<{\centering} | p{0.78cm}<{\centering} p{0.82cm}<{\centering} | p{0.78cm}<{\centering}  p{0.82cm}<{\centering} | p{0.78cm}<{\centering} p{0.82cm}<{\centering} | p{0.78cm}<{\centering} p{0.82cm}<{\centering} | p{0.78cm}<{\centering} p{0.82cm}<{\centering} | p{0.78cm}<{\centering} p{0.82cm}<{\centering}}
\toprule
\multirow{2}*{Algorithms} & \multicolumn{2}{c}{Ratio=0.01} & \multicolumn{2}{c}{Ratio=0.10} & \multicolumn{2}{c}{Ratio=0.25} & \multicolumn{2}{c}{Ratio=0.30} & \multicolumn{2}{c}{Ratio=0.40} & \multicolumn{2}{c}{Avg.}\\
\cline{2-13}
&PSNR&SSIM&PSNR&SSIM&PSNR&SSIM&PSNR&SSIM&PSNR&SSIM&PSNR&SSIM\\

\hline

\footnotesize{CSNet}\tiny{${\rm \textcolor{blue}{(ICME2017)}}$}\footnotesize{\cite{8019428}}&24.02&0.6378&32.30&0.9015&36.63&0.9562&37.90&0.9630&39.89&0.9736&34.15&0.8864\\
\footnotesize{LapCSNet}\tiny{${\rm \textcolor{blue}{(ICASSP2018)}}$}\footnotesize{\cite{cui2018efficient}}&24.42&0.6686&32.44&0.9047&--\ --&--\ --&--\ --&--\ --&--\ --&--\ --&--\ --&--\ --\\
\footnotesize{SCSNet}\tiny{${\rm \textcolor{blue}{(CVPR2019)}}$}\footnotesize{\cite{shi2019scalable}}&24.21&0.6468&32.77&0.9083&37.20&0.9558&38.45&0.9655&40.44&0.9755&34.61&0.8904\\
\footnotesize{CSNet$^{+}$}\tiny{${\rm \textcolor{blue}{(TIP2020)}}$}\footnotesize{\cite{8765626}}&24.18&0.6478&32.59&0.9062&37.11&0.9560&38.25&0.9644&40.11&0.9740&34.45&0.8897\\
\footnotesize{NL-CSNet}\tiny{${\rm \textcolor{blue}{(TMM2021)}}$}\footnotesize{\cite{9635679}}&\textbf{24.82}&\textbf{0.6771}&33.84&\underline{0.9312}&37.78&0.9635&38.86&0.9703&40.69&0.9778&35.20&\underline{0.9040}\\
\hline
\footnotesize{BCS-Net}\tiny{${\rm \textcolor{blue}{(TMM2020)}}$}\footnotesize{\cite{9159912}}&22.98&0.6103&32.71&0.9030&37.90&0.9576&38.64&0.9694&39.88&0.9785&34.42&0.8838\\
\footnotesize{OPINENet$^{+}$}\tiny{${\rm \textcolor{blue}{(JSTSP2020)}}$}\footnotesize{\cite{9019857}}&22.76&0.6194&33.72&0.9259&38.05&0.9631&39.14&0.9689&41.07&0.9780&34.95&0.8911\\
\footnotesize{AMP-Net$^{+}$}\tiny{${\rm \textcolor{blue}{(TIP2021)}}$}\footnotesize{\cite{9298950}}&23.00&0.6488&33.35&0.9162&38.00&0.9606&39.07&0.9671&40.94&0.9765&34.87&0.8938\\
\footnotesize{COAST}\tiny{${\rm \textcolor{blue}{(TIP2021)}}$}\footnotesize{\cite{9467810}}&23.31&0.6514&\underline{33.90}&0.9266&38.21&0.9648&39.23&0.9706&41.36&0.9780&35.20&0.8983\\
\footnotesize{MADUN}\tiny{${\rm \textcolor{blue}{(ACMMM2021)}}$}\footnotesize{\cite{2021Memory}}&23.12&0.6503&33.86&0.9267&\underline{38.44}&\underline{0.9660}&\underline{39.57}&\underline{0.9723}&\underline{41.72}&\underline{0.9808}&\underline{35.34}&0.8992\\
\midrule

FHDUN&\underline{24.04}&\underline{0.6705}&\textbf{34.25}&\textbf{0.9345}&\textbf{38.78}&\textbf{0.9682}&\textbf{39.90}&\textbf{0.9734}&\textbf{41.98}&\textbf{0.9812}&\textbf{35.79}&\textbf{0.9056}\\
\bottomrule

\end{tabular}
\vspace{-0.07in}
\label{tab:4}
\end{table*}

\begin{table*}[t]
\centering
\caption{Average PSNR and SSIM comparisons of recent deep network-based CS algorithms using learned sampling matrix on dataset Set14. Bold indicates the best result, underline signifies the second best result.}
\label{tab:14}
\vspace{-0.15in}
\small
\begin{tabular}{p{3.5cm}<{\centering} | p{0.78cm}<{\centering} p{0.82cm}<{\centering} | p{0.78cm}<{\centering}  p{0.82cm}<{\centering} | p{0.78cm}<{\centering} p{0.82cm}<{\centering} | p{0.78cm}<{\centering} p{0.82cm}<{\centering} | p{0.78cm}<{\centering} p{0.82cm}<{\centering} | p{0.78cm}<{\centering} p{0.82cm}<{\centering}}
\toprule
\multirow{2}*{Algorithms} & \multicolumn{2}{c}{Ratio=0.01} & \multicolumn{2}{c}{Ratio=0.10} & \multicolumn{2}{c}{Ratio=0.25} & \multicolumn{2}{c}{Ratio=0.30} & \multicolumn{2}{c}{Ratio=0.40} & \multicolumn{2}{c}{Avg.}\\
\cline{2-13}
&PSNR&SSIM&PSNR&SSIM&PSNR&SSIM&PSNR&SSIM&PSNR&SSIM&PSNR&SSIM\\

\hline

\footnotesize{CSNet}\tiny{${\rm \textcolor{blue}{(ICME2017)}}$} \footnotesize{\cite{8019428}}&22.79&0.5628&28.91&0.8119&32.86&0.9057&34.00&0.9276&35.84&0.9481&30.88&0.8312\\
\footnotesize{LapCSNet}\tiny{${\rm \textcolor{blue}{(ICASSP2018)}}$}\footnotesize{\cite{cui2018efficient}}&23.16&0.5818&29.00&0.8147&--\ --&--\ --&--\ --&--\ --&--\ --&--\ --&--\ --&--\ --\\
\footnotesize{SCSNet}\tiny{${\rm \textcolor{blue}{(CVPR2019)}}$}\footnotesize{\cite{shi2019scalable}}&22.87&0.5631&29.22&0.8181&33.24&0.9073&34.51&0.9311&36.54&0.9525&31.28&0.8344\\
\footnotesize{CSNet$^{+}$}\tiny{${\rm \textcolor{blue}{(TIP2020)}}$}\footnotesize{\cite{8765626}}&22.83&0.5630&29.13&0.8169&33.19&0.9064&34.34&0.9297&36.16&0.9502&31.13&0.8332\\
\footnotesize{NL-CSNet}\tiny{${\rm \textcolor{blue}{(TMM2021)}}$}\footnotesize{\cite{9635679}}&\textbf{23.61}&\underline{0.5862}&30.16&\underline{0.8527}&33.84&0.9270&34.88&0.9405&36.86&0.9573&31.87&\underline{0.8527}\\
\hline
\footnotesize{BCS-Net}\tiny{${\rm \textcolor{blue}{(TMM2020)}}$}\footnotesize{\cite{9159912}}&22.38&0.5543&29.73&0.8384&34.50&0.9279&35.53&0.9390&37.46&0.9537&31.92&0.8427\\
\footnotesize{OPINENet$^{+}$}\tiny{${\rm \textcolor{blue}{(JSTSP2020)}}$}\footnotesize{\cite{9019857}}&22.30&0.5508&29.94&0.8415&34.31&0.9268&35.18&0.9369&37.51&0.9572&31.85&0.8426\\
\footnotesize{AMP-Net$^{+}$}\tiny{${\rm \textcolor{blue}{(TIP2021)}}$}\footnotesize{\cite{9298950}}&22.60&0.5723&29.87&0.8130&34.27&0.9218&35.23&0.9364&37.42&0.9561&31.88&0.8399\\
\footnotesize{COAST}\tiny{${\rm \textcolor{blue}{(TIP2021)}}$}\footnotesize{\cite{9467810}}&22.81&0.5764&\underline{30.26}&0.8507&34.72&0.9335&35.66&0.9404&37.86&0.9598&32.26&0.8522\\
\footnotesize{MADUN}\tiny{${\rm \textcolor{blue}{(ACMMM2021)}}$}\footnotesize{\cite{2021Memory}}&22.44&0.5675&30.17&0.8483&\underline{34.98}&\underline{0.9362}&\underline{36.03}&\underline{0.9473}&\underline{38.27}&\underline{0.9641}&\underline{32.38}&\underline{0.8527}\\
\midrule

FHDUN&\underline{23.23}&\textbf{0.5906}&\textbf{30.76}&\textbf{0.8596}&\textbf{35.32}&\textbf{0.9381}&\textbf{36.41}&\textbf{0.9489}&\textbf{38.55}&\textbf{0.9645}&\textbf{32.85}&\textbf{0.8603}\\
\bottomrule

\end{tabular}
\vspace{-0.13in}
\end{table*}

\section{Experiments and Analysis}


\subsection{Loss Function}

Given the full-sampled image $x_{i}$ and the sampling matrix $\Phi$, the compressed measurements can be obtained by $y_{i}=\Phi x_{i}$. Our proposed FHDUN takes $y_{i}$ and $\Phi$ as inputs and aims to narrow down the gap between the output and the target image $x_{i}$. Since the proposed FHDUN consists of multiple phases, series of outputs $\{(x_{i})_{t}^{k}\}$ are generated through the pipeline of the entire framework, where $t$ represents the scale factor in the hierarchical structure, and $k$ is the index of the phase number in the proposed CS framework.

In our proposed FHDUN, the outputs of different scales in all phases are constrained. Specifically, we directly use the L2 norm to restrain the distance between the outputs $(x_{i})_{t}^{k}$ and the corresponding real entities $S_{t}(x_{i})$, i.e.,

\vspace{-0.15in}
\begin{equation}
\mathcal{L}(\Theta) = \frac{1}{KN_{a}}\sum_{i=1}^{N_{a}}\sum_{k=1}^{K}\sum_{t\in T}||(x_{i})_{t}^{k} - S_{t}(x_{i})||_{F}^{2}
\end{equation}
where $\Theta$ denotes the learnable parameter set of our proposed FHDUN, including $\{\theta_{u}^{k}\}$ in MBAM, $\{\theta_{r}^{k}\}$ in AGDM and $\{\theta_{x}^{k}\}$ in HPMM. It worth noting that since multiple entities of different scales are revealed during the network execution, we use $S_{t}(x_{i})$, rather than $x_{i}$, as the corresponding label. In addition, $N_{a}$, $T$ and $K$ represent the number of training images, scale factor set in the hierarchical structure and the phase number of our FHDUN respectively.

\vspace{-0.05in}
\subsection{Implementation and Training Details}

In the proposed CS framework, we set the scale factor set $T$=$\{1,2,4\}$, that is to say, the proposed hierarchical framework consists of three horizontal branches of different scales (as shown in Fig.1). Considering the phase number $K$, Fig.6 shows the relationship between the phase number and the reconstructed quality, and we set $K$=8 in our model, which saves more than 50\% iteration loops against recent DUNs (such as MADUN\cite{2021Memory} and COAST\cite{9467810}). For more configuration details of networks HPG-Net and HPM-Net in our framework, the channel numbers of the intermediate feature maps in their three horizontal branches are respectively set as 16, 32 and 64. In the residual blocks of HPM-Net, the number of convolutional layer is set as $d$=3. In addition, with the exception of the convolutional layer in the feature aggregation submodule (in Fig.5(b)), which has the convolutional kernel size of 1$\times$1, the kernel size of all other convolutional layers is set to 3$\times$3. We initialize the convolutional filters using the same method as\cite{he2015delving} and pad zeros around the boundaries to keep the size of all feature maps.

\begin{table*}[t]
\centering
\caption{Average PSNR and SSIM comparisons of recent deep unfolding CS networks using learned sampling matrix on dataset BSD68. Bold indicates the best result, underline signifies the second best result.}
\label{tab:5}
\vspace{-0.15in}
\small
\begin{tabular}{p{3.5cm}<{\centering} | p{0.78cm}<{\centering} p{0.82cm}<{\centering} | p{0.78cm}<{\centering}  p{0.82cm}<{\centering} | p{0.78cm}<{\centering} p{0.82cm}<{\centering} | p{0.78cm}<{\centering} p{0.82cm}<{\centering} | p{0.78cm}<{\centering} p{0.82cm}<{\centering} | p{0.78cm}<{\centering} p{0.82cm}<{\centering}}
\toprule
\multirow{2}*{Algorithms} & \multicolumn{2}{c}{Ratio=0.01} & \multicolumn{2}{c}{Ratio=0.10} & \multicolumn{2}{c}{Ratio=0.25} & \multicolumn{2}{c}{Ratio=0.30} & \multicolumn{2}{c}{Ratio=0.40} & \multicolumn{2}{c}{Avg.}\\
\cline{2-13}
&PSNR&SSIM&PSNR&SSIM&PSNR&SSIM&PSNR&SSIM&PSNR&SSIM&PSNR&SSIM\\

\hline

\footnotesize{BCS-Net}\tiny{${\rm \textcolor{blue}{(TMM2020)}}$}\footnotesize{\cite{9159912}}&22.16&0.5287&27.78&0.7864&31.14&0.9006&32.15&0.9167&33.90&0.9473&29.43&0.8159\\
\footnotesize{OPINENet$^{+}$}\tiny{${\rm \textcolor{blue}{(JSTSP2020)}}$}\footnotesize{\cite{9019857}}&21.88&0.5162&27.81&0.8040&31.50&0.9062&32.78&0.9278&34.73&0.9521&29.74&0.8213\\
\footnotesize{AMP-Net$^{+}$}\tiny{${\rm \textcolor{blue}{(TIP2021)}}$}\footnotesize{\cite{9298950}}&21.94&0.5253&\underline{27.86}&0.7928&31.75&0.9050&32.84&0.9242&34.86&0.9509&29.85&0.8196\\
\footnotesize{COAST}\tiny{${\rm \textcolor{blue}{(TIP2021)}}$}\footnotesize{\cite{9467810}}&\underline{22.30}&\underline{0.5391}&27.80&0.8091&31.81&0.9128&32.78&0.9331&34.90&0.9565&\underline{29.92}&\underline{0.8301}\\
\footnotesize{MADUN}\tiny{${\rm \textcolor{blue}{(ACMMM2021)}}$}\footnotesize{\cite{2021Memory}}&21.65&0.5249&27.74&\underline{0.8108}&\underline{31.90}&\underline{0.9165}&\underline{32.96}&\underline{0.9353}&\underline{35.02}&\underline{0.9584}&29.86&0.8293\\
\midrule

FHDUN&\textbf{22.58}&\textbf{0.5395}&\textbf{28.29}&\textbf{0.8173}&\textbf{32.16}&\textbf{0.9189}&\textbf{33.24}&\textbf{0.9360}&\textbf{35.33}&\textbf{0.9589}&\textbf{30.32}&\textbf{0.8341}\\
\bottomrule

\end{tabular}
\vspace{-0.1in}
\label{tab:6}
\end{table*}

\begin{table*}[t]
\centering
\caption{Average PSNR and SSIM comparisons of recent deep unfolding CS networks using learned sampling matrix on dataset Urban100. Bold indicates the best result, underline signifies the second best result.}
\label{tab:7}
\vspace{-0.15in}
\small
\begin{tabular}{p{3.5cm}<{\centering} | p{0.78cm}<{\centering} p{0.82cm}<{\centering} | p{0.78cm}<{\centering}  p{0.82cm}<{\centering} | p{0.78cm}<{\centering} p{0.82cm}<{\centering} | p{0.78cm}<{\centering} p{0.82cm}<{\centering} | p{0.78cm}<{\centering} p{0.82cm}<{\centering} | p{0.78cm}<{\centering} p{0.82cm}<{\centering}}
\toprule
\multirow{2}*{Algorithms} & \multicolumn{2}{c}{Ratio=0.01} & \multicolumn{2}{c}{Ratio=0.10} & \multicolumn{2}{c}{Ratio=0.25} & \multicolumn{2}{c}{Ratio=0.30} & \multicolumn{2}{c}{Ratio=0.40} & \multicolumn{2}{c}{Avg.}\\
\cline{2-13}
&PSNR&SSIM&PSNR&SSIM&PSNR&SSIM&PSNR&SSIM&PSNR&SSIM&PSNR&SSIM\\

\hline

\footnotesize{BCS-Net}\tiny{${\rm \textcolor{blue}{(TMM2020)}}$}\footnotesize{\cite{9159912}}&20.64&0.5148&28.26&0.8469&33.52&0.9348&34.51&0.9532&36.76&0.9680&30.74&0.8435\\
\footnotesize{OPINENet$^{+}$}\tiny{${\rm \textcolor{blue}{(JSTSP2020)}}$}\footnotesize{\cite{9019857}}&20.56&0.5140&28.84&0.8675&33.77&0.9424&34.91&0.9557&36.98&0.9708&31.01&0.8502\\
\footnotesize{AMP-Net$^{+}$}\tiny{${\rm \textcolor{blue}{(TIP2021)}}$}\footnotesize{\cite{9298950}}&\underline{20.94}&\underline{0.5386}&28.80&0.8589&33.76&0.9383&34.88&0.9546&36.92&0.9701&31.06&0.8521\\
\footnotesize{COAST}\tiny{${\rm \textcolor{blue}{(TIP2021)}}$}\footnotesize{\cite{9467810}}&20.76&0.5279&28.82&0.8690&33.81&0.9452&34.96&0.9573&37.10&0.9712&31.09&0.8540\\
\footnotesize{MADUN}\tiny{${\rm \textcolor{blue}{(ACMMM2021)}}$}\footnotesize{\cite{2021Memory}}&20.43&0.5250&\underline{28.86}&\underline{0.8692}&\underline{33.98}&\underline{0.9467}&\underline{35.20}&\underline{0.9588}&\underline{37.35}&\underline{0.9723}&\underline{31.17}&\underline{0.8544}\\
\midrule

FHDUN&\textbf{21.18}&\textbf{0.5434}&\textbf{29.20}&\textbf{0.8763}&\textbf{34.25}&\textbf{0.9486}&\textbf{35.48}&\textbf{0.9597}&\textbf{37.55}&\textbf{0.9726}&\textbf{31.53}&\textbf{0.8601}\\
\bottomrule

\end{tabular}
\vspace{-0.12in}
\label{tab:8}
\end{table*}

We use the training set (400 images) from the BSD500\cite{Arbelaez2011Contour} dataset as our training data. Furthermore, we augment the training data in the following two ways: ($1$) Rotate the images by 90$^{\circ}$, 180$^{\circ}$ and 270$^{\circ}$ randomly. ($2$) Flip the images horizontally with a probability of 0.5. In the training process, we set block size as 32, i.e., $N$=1024, and in order to alleviate blocking artifacts, we randomly crop the size of patches to 96$\times$96. More specifically, we first unfold the blocks of size 96$\times$96 into non-overlapping blocks of size 32$\times$32 in the sampling process and then concatenate all the blocks together in the initial reconstruction. We also unfold the whole testing image with this approach during testing process. We use the PyTorch toolbox and train our model using the Adaptive moment estimation (Adam) solver on a NVIDIA GTX 3090 GPU. Furthermore, we set the momentum to 0.9 and the weight decay to 1e-4. The learning rate is initialized to 1e-4 for all layers and decreased by a factor of 2 for every 30 epochs. We train our model for 200 epochs totally and 1000 iterations are performed for each epoch. Therefore, 200$\times$1000 iterations are completed in the whole training process.

\vspace{-0.07in}
\subsection{Comparisons with Other Methods}
Depending on whether the sampling matrix $\Phi$ can be learned jointly with the reconstruction process, we conduct two types of experimental comparisons: comparisons with the random sampling matrix-based CS methods and the learned sampling matrix-based CS methods. For testing data, we carry out extensive experiments on several representative benchmark datasets: Set5\cite{shi2019scalable}, Set14\cite{cui2018efficient}, Set11\cite{eccvcs}, BSD68\cite{9199540} and Urban100\cite{2021Memory}, which are widely used in the recent CS-related works. To ensure the fairness, we evaluate the reconstruction performance with two widely used quality evaluation metrics: PSNR and SSIM in terms of various sampling ratios.

\textbf{Random Sampling Matrix:} For the random sampling matrix-based CS methods, we compare our proposed FHDUN with eleven recent representative deep network-based reconstruction algorithms, including four deep black box networks (ReconNet\cite{7780424}, I-Recon\cite{2018Convolutional}, DR$^{2}$-Net\cite{Yao2017DR2} and DPA-Net\cite{9199540}) and seven DUNs (IRCNN\cite{8099783}, LD-AMP \cite{2017Learned}, ISTA-Net$^{+}$\cite{8578294}, DPDNN\cite{8481558}, NN\cite{8878159}, MAC-Net\cite{eccvcs} and iPiano-Net\cite{SU2020115989}). For these compared methods, we train them with the same experimental configurations as\cite{2021Memory}. In our FHDUN, the orthogonalized Gaussian random matrix\cite{9199540,8999514} is utilized. Table 1 presents the average PSNR and SSIM comparisons at five sampling ratios (i.e., 0.01, 0.10, 0.25, 0.30, 0.40) on dataset Set11, from which we can observe that the proposed FHDUN outperforms all the compared methods in PSNR and SSIM by large margins. 

\begin{figure}[h]
\centering
\vspace{-0.1in}
\includegraphics[width=3.2in]{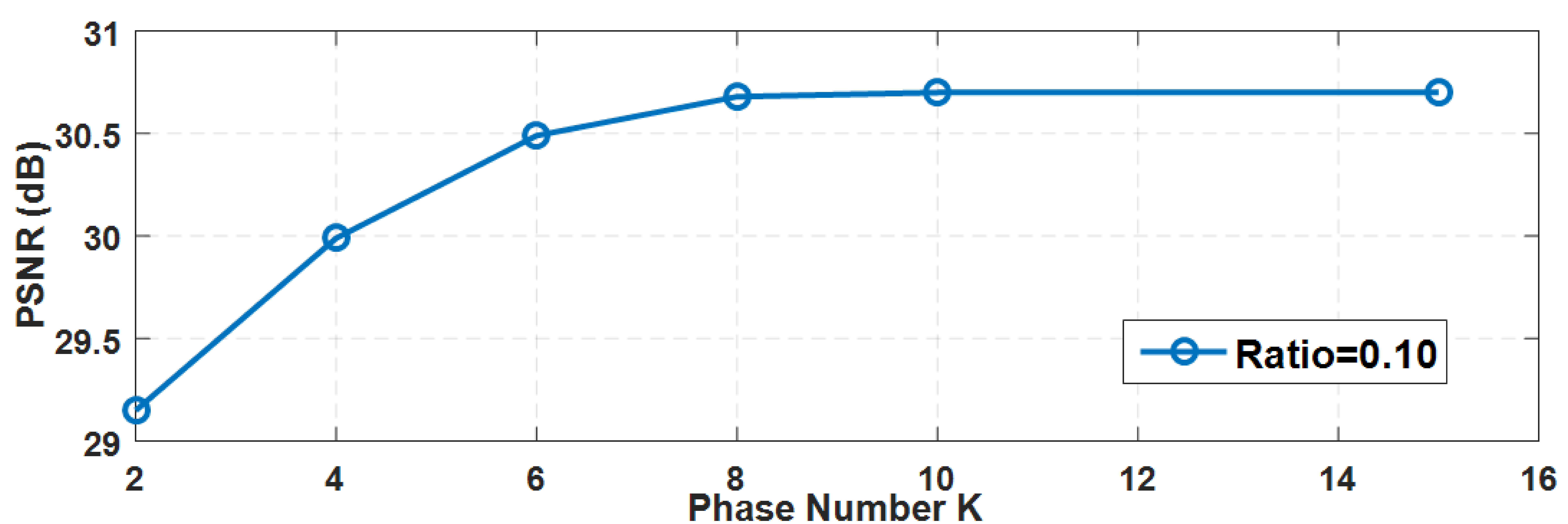}
\vspace{-0.1in}\caption{The relationship between the phase number and the image quality (PSNR) on dataset Set11.}
\vskip -0.2in
\label{Fig:6}
\end{figure}

\textbf{Learned Sampling Matrix:} For the learned sampling matrix-based CS methods, the sampling matrix is jointly optimized with the reconstruction module, which facilitates the collaborations between the sampling and reconstruction. For comparative fairness, ten recent deep network-based literatures, i.e., CSNet\cite{8019428}, LapCSNet\cite{cui2018efficient}, SCSNet\cite{shi2019scalable}, CSNet$^{+}$\cite{8765626}, NL-CSNet\cite{9635679}, BCS-Net\cite{9159912}, OPINE-Net \cite{9019857}, AMP-Net\cite{9298950}, COAST\cite{9467810} and MADUN\cite{2021Memory} participate in the comparison in our experiments, and the first five algorithms are the black box CS networks, the last five schemes belong to the deep unfolding models. For these compared methods, we obtain the experimental results by running their published source codes. Tables 2,3 present the experimental comparisons at the given five sampling ratios on dataset Set5 and Set14, from which we can observe that the proposed FHDUN performs much better than the other deep network-based CS schemes. To further evaluate the performance of our proposed FHDUN, we conduct more comparisons compared with recent CS DUNs on two benchmark datasets BSD68 and Urban100, and the results are shown in Tables 4, 5. Compared with the recent MADUN, the proposed FHDUN can achieve on average 0.93dB, 0.55dB, 0.26dB, 0. 28dB, 0.31dB gains on dataset BSD68. For dataset Urban100, the proposed FHDUN achieves on average 0.75dB, 0.34dB, 0.27dB, 0.28dB, 0.20dB gains in PSNR compared with MADUN. The visual comparisons are shown in Fig. 7, from which we observe that the proposed method can preserve more details and retain sharper edges compared to the other CS networks.

\begin{table}[b]
\vspace{-0.23in}
  \caption{The ablation results of the proposed FHDUN on dataset Set14.}
  \label{tab:9}
  \vspace{-0.15in}

  \begin{tabular}{p{0.84cm}<{\centering}  p{0.84cm}<{\centering}  p{0.84cm}<{\centering}  | p{0.87cm}<{\centering}   p{0.87cm}<{\centering}  p{0.87cm}<{\centering}  p{0.87cm}<{\centering}}
    \toprule

    \footnotesize{MBAM}&\footnotesize{AGDM}&\footnotesize{HPMM}&\small{0.10}&\small{0.25}&\small{0.30}&\small{0.40}\\
    \midrule
    \scriptsize{\XSolidBrush} & \checkmark & \checkmark & 30.59 & 35.22 & 36.24 & 38.42\\

    $\checkmark $ & \scriptsize{\XSolidBrush} & $\checkmark $ & 30.51 & 35.09 & 36.16 & 38.28\\

    $\checkmark $ & $\checkmark $ & \scriptsize{\XSolidBrush} & 30.18 & 34.68 & 35.95 & 38.17\\
    \midrule
    \checkmark & $\checkmark $ &$\checkmark $ & \textbf{30.76} &  \textbf{35.32} & \textbf{36.41}& \textbf{38.55}\\
  \bottomrule
\end{tabular}
\end{table}

\begin{figure*}[t]
\hspace{-0.08in}
\begin{minipage}[t]{0.133\textwidth}
\centering
\includegraphics[width=0.97in]{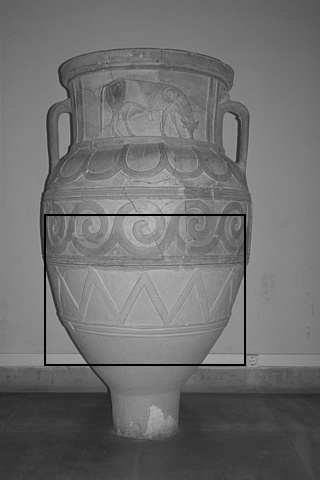}
\begin{scriptsize}
\centering
\vskip -0.46 cm \begin{tiny}GT$\backslash$PSNR$\backslash$SSIM\end{tiny}
\end{scriptsize}
\end{minipage}
\hspace{-0.012in}
\begin{minipage}[t]{0.133\textwidth}
\centering
\includegraphics[width=0.97in]{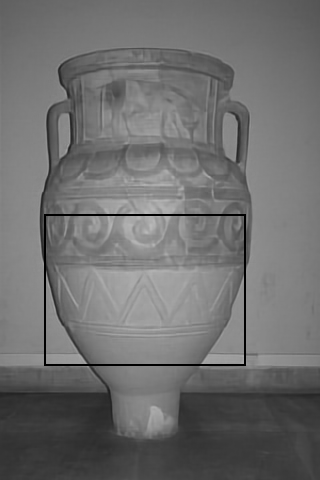}
\begin{scriptsize}
\centering
\vskip -0.46 cm \begin{tiny}NLCSNet$\backslash$35.44$\backslash$0.9098\end{tiny}
\end{scriptsize}
\end{minipage}
\hspace{-0.012in}
\begin{minipage}[t]{0.133\textwidth}
\centering
\includegraphics[width=0.97in]{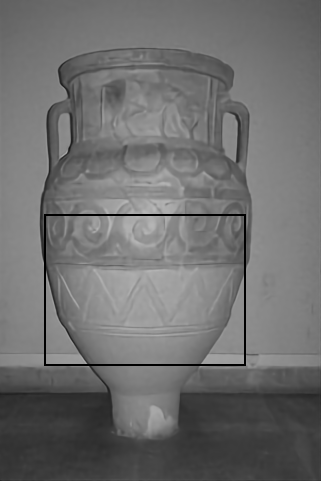}
\begin{scriptsize}
\centering
\vskip -0.46 cm \begin{tiny}OPINE$\backslash$35.24$\backslash$0.9021\end{tiny}
\end{scriptsize}
\end{minipage}
\hspace{-0.012in}
\begin{minipage}[t]{0.133\textwidth}
\centering
\includegraphics[width=0.97in]{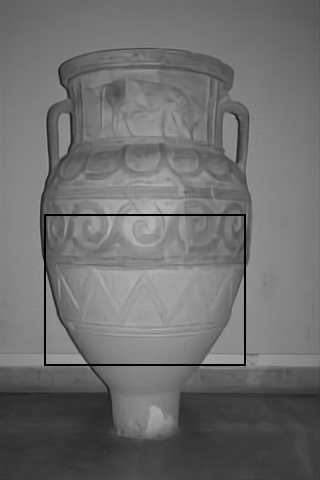}
\begin{scriptsize}
\centering
\vskip -0.46 cm \begin{tiny}AMPNet$\backslash$34.85$\backslash$0.8900\end{tiny}
\end{scriptsize}
\end{minipage}
\hspace{-0.012in}
\begin{minipage}[t]{0.133\textwidth}
\centering
\includegraphics[width=0.97in]{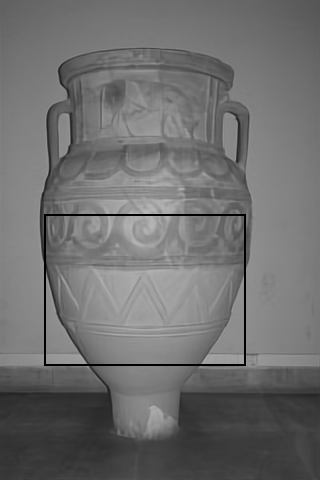}
\begin{scriptsize}
\centering
\vskip -0.46 cm \begin{tiny}COAST$\backslash$35.60$\backslash$0.9104\end{tiny}
\end{scriptsize}
\end{minipage}
\hspace{-0.012in}
\begin{minipage}[t]{0.133\textwidth}
\centering
\includegraphics[width=0.97in]{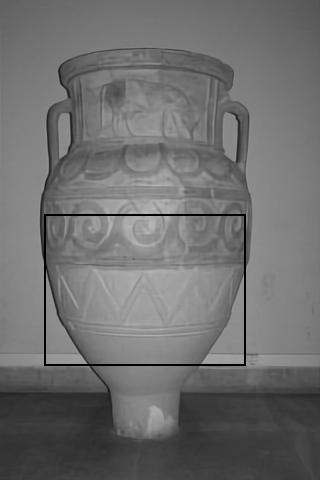}
\begin{scriptsize}
\centering
\vskip -0.46 cm \begin{tiny}MADUN$\backslash$35.58$\backslash$0.9114\end{tiny}
\end{scriptsize}
\end{minipage}
\hspace{-0.012in}
\begin{minipage}[t]{0.133\textwidth}
\centering
\includegraphics[width=0.97in]{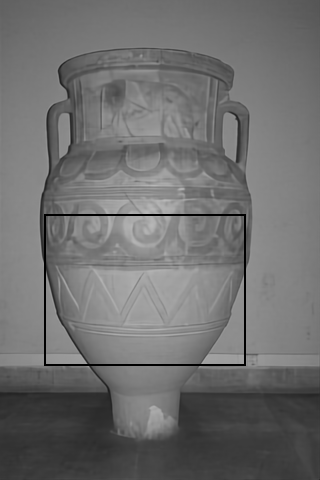}
\begin{scriptsize}
\centering
\vskip -0.46 cm \begin{tiny}Ours$\backslash$\textbf{36.15}$\backslash$\textbf{0.9210}\end{tiny}
\end{scriptsize}
\end{minipage}

\label{fig:10}
\vspace{-0.16in}
\end{figure*}

\begin{figure*}[t]
\hspace{-0.08in}
\begin{minipage}[t]{0.133\textwidth}
\centering
\includegraphics[width=0.97in]{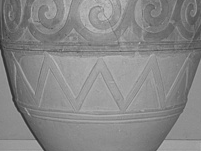}
\begin{scriptsize}
\end{scriptsize}
\end{minipage}
\hspace{-0.012in}
\begin{minipage}[t]{0.133\textwidth}
\centering
\includegraphics[width=0.97in]{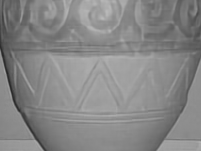}
\begin{scriptsize}
\end{scriptsize}
\end{minipage}
\hspace{-0.012in}
\begin{minipage}[t]{0.133\textwidth}
\centering
\includegraphics[width=0.97in]{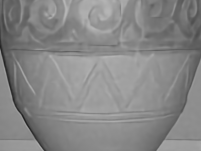}
\begin{scriptsize}
\end{scriptsize}
\end{minipage}
\hspace{-0.012in}
\begin{minipage}[t]{0.133\textwidth}
\centering
\includegraphics[width=0.97in]{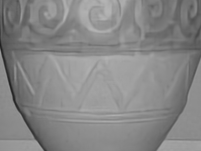}
\begin{scriptsize}
\end{scriptsize}
\end{minipage}
\hspace{-0.012in}
\begin{minipage}[t]{0.133\textwidth}
\centering
\includegraphics[width=0.97in]{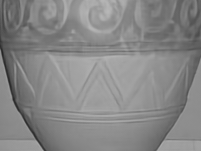}
\begin{scriptsize}
\end{scriptsize}
\end{minipage}
\hspace{-0.012in}
\begin{minipage}[t]{0.133\textwidth}
\centering
\includegraphics[width=0.97in]{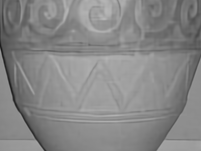}
\begin{scriptsize}
\end{scriptsize}
\end{minipage}
\hspace{-0.012in}
\begin{minipage}[t]{0.133\textwidth}
\centering
\includegraphics[width=0.97in]{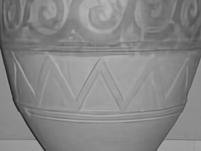}
\begin{scriptsize}
\end{scriptsize}
\end{minipage}

\vspace{-0.25in}
\caption{Visual quality comparisons between the proposed FHDUN and recent deep network-based CS methods that use learned sampling matrix on one testing image sample from BSD68 at sampling ratio of 0.10.}
\vspace{-0.06in}
\label{fig:11}
\vspace{-0.1in}
\end{figure*}


From Tables 2,3, we observe that NL-CSNet\cite{9635679} outperforms FHDUN at lowest sampling rate of 0.01. A reasonable explanation is provided bellow. NL-CSNet is a deep black box CS network, which directly build an inverse deep mapping from the measurement domain to the image domain. Different from DUNs, the sampling matrix of NL-CSNet is not well embedded into its reconstruction process, which leads to the sampling matrix cannot provide the necessary guidance for the image reconstruction, thus making the reconstructed performance of\cite{9635679} seriously depends on the network structure design and its careful tuning strategy. Apparently, with the increase of sampling ratio, the dimension of sampling matrix is higher, which can provide more guidance for the reconstruction. Therefore, with the sampling ratio increases, the proposed FHDUN achieves much better reconstruction performance against NL-CSNet. Specifically, on the one hand, when sampling ratio is very low (such as 0.01), the guidance provided by the sampling matrix is very limited, and the reconstruction performance mainly comes from the network design and fine-tuning policy. As above, the complexed network structure of NL-CSNet makes better reconstructed quality compared with our FHDUN at low sampling ratio 0.01. On the other hand, when the sampling ratio becomes larger, the sampling matrix can provide more guidance, and therefore the proposed FHDUN begins to obtain better reconstruction quality against NL-CSNet. This phenomenon is also reflected in other CS methods as shown in Tables 2,3.

Considering the parameter quantities and running speed. Because MADUN\cite{2021Memory} achieves the best performance in the compared methods, we mainly analyze the comparisons against MADUN. The following analysis is based on the models with the sampling ratio of 0.10. For the number of parameters, the parameter quantities of MADUN and FHDUN respectively are 3.02M and 3.68M. For running speed, we test three images with different resolutions of 256x256, 720x576, 1024x768 in the same platform (GTX 3090 GPU), and the running times (in second) of MADUN and FHDUN are 0.0364, 0.0867, 0.2132 and 0.0349, 0.0786, 0.1980 respectively. As above, due to the hierarchical architecture, the proposed FHUDN has more parameters compared with MADUN. However, because most of the convolutional layers are performed on lower resolution feature maps, the running speed of the proposed FHDUN is competitive with MADUN.

\vspace{-0.08in}
\subsection{Ablation Studies}

As above, the proposed CS framework achieves higher reconstruction quality. In order to evaluate the contributions of each part of the proposed FHDUN, we design several variations of the proposed model, in which some functional modules are selectively discarded or replaced. Table 6 shows the experimental results on the dataset Set14, in which three functional modules, i.e., MBAM, AGDM and HPMM, are considered. It should be noted that when MBAM or AGDM are discarded, we replace our HPG-Net with the hyperparametric learning method in the existing DUNs\cite{8578294,9019857}. Besides, when HPMM is discarded, we only retain the first horizontal branch of the proposed hierarchical architecture for CS reconstruction. The experimental results reveal that the proposed hierarchical network architecture can bring a maximum gain compared with other two modules. As a note, each of these three modules can enhance the CS reconstruction performance to varying degrees. 

To further verify the effectiveness of the introduced acceleration module MBAM, a new variant without MBAM module, dubbed FHDUN(w/o-M), is designed. We train this variant from scratch with the same configurations as FHDUN, and we find that the convergence of FHDUN(w/o-M) needs about 15 iterations (testing on dataset Set11 at sampling ratio of 0.10 same as Fig.6). Compared with the recent MADUN\cite{2021Memory} and COAST\cite{9467810}, the variant FHDUN(w/o-M) clearly requires fewer iterations, which may be caused by the newly introduced network components, such as hierarchical architecture and hyperparametric generation network. On the other hand, considering the iterations of FHDUN(w/o-M) and FHDUN, we can get that the introduced acceleration module MBAM can save nearly half of the iterations.

\vspace{-0.08in}
\section{Conclusion}
In this paper, a novel Fast Hierarchical Deep Unfolding CS Network (FHDUN) based on the traditional solver FISTA is proposed, in which a well-designed hierarchical architecture is developed to cooperatively perceive the richer contextual prior information in multi-scale spaces. Due to the hierarchical design, the proposed framework is able to persist and transmit richer textures and structures among cascading iterations for boosting reconstructed quality. In addition, series of hyperparametric generation networks are developed in our framework to dynamically produce the corresponding hyperparameters according to the input content, which enhances the adaptability of the proposed model. Moreover, by unfolding the traditional FISTA, the newly introduced acceleration module can save more than 50\% of the iterative loops against recent DUNs. Extensive CS experiments manifest that the proposed FHDUN outperforms existing state-of-the-art CS methods.

\section{ACKNOWLEDGMENTS}
This work was supported by the National Natural Science Foundation of China under Grant 61872116.


\bibliographystyle{ACM-Reference-Format}
\balance
\bibliography{sample-base}


\end{document}